\documentclass[11pt]{article}

\usepackage[preprint]{acl}

\usepackage{amsmath}
\usepackage{amsfonts}
\usepackage{booktabs,multirow}
\usepackage{makecell}
\usepackage{bm}
\usepackage{adjustbox}
\usepackage[table]{xcolor}
\usepackage{subcaption}
\usepackage{multicol}
\usepackage{enumitem}
\usepackage{tabularx}
\usepackage{array}

\usepackage{graphicx}
\usepackage{subcaption}
\usepackage{algorithm}
\usepackage{algpseudocode}

\algnewcommand{\Phase}[1]{%
  \Statex \hspace{-\algorithmicindent}{\footnotesize\textsc{#1}}%
}

\usepackage{tcolorbox}
\tcbuselibrary{breakable,skins}

\newtcolorbox{promptbox}[1][]{
  colback=gray!10,      
  colframe=gray!60,     
  coltitle=black,       
  fonttitle=\bfseries,  
  boxrule=0.5mm,        
  arc=2mm,              
  title={#1}            
}

\usepackage[utf8]{inputenc}
\usepackage{geometry}
\usepackage{xcolor}
\usepackage{tcolorbox}
\usepackage{enumitem}

\usepackage{tikz}

\usepackage{booktabs}
\usepackage{multirow}
\usepackage{colortbl} 
\definecolor{graybg}{rgb}{0.95, 0.95, 0.95}

\usepackage{times}
\usepackage{latexsym}

\usepackage[T1]{fontenc}

\usepackage[utf8]{inputenc}

\usepackage{microtype}

\usepackage{inconsolata}

\title{\textsc{EvoTool}: Self-Evolving Tool-Use Policy Optimization in LLM Agents via Blame-Aware Mutation and Diversity-Aware Selection}

\author{Shuo Yang,
Soyeon Caren Han\thanks{Soyeon Caren Han is the corresponding author},
Xueqi Ma, 
Yan Li, \\
{\bf Mohammad Reza Ghasemi Madani },
{\bf Eduard Hovy} \\
School of Computing and Information Systems, The University of Melbourne \\
}

\begin{document}
\maketitle
\begin{abstract}
LLM-based agents depend on effective tool-use policies to solve complex tasks, yet optimizing these policies remains challenging due to delayed supervision and the difficulty of credit assignment in long-horizon trajectories. Existing optimization approaches tend to be either monolithic, which are prone to entangling behaviors, or single-aspect, which ignore cross-module error propagation. To address these limitations, we propose \textsc{EvoTool}, a self-evolving framework that optimizes a modular tool-use policy via a gradient-free evolutionary paradigm.
\textsc{EvoTool} decomposes agent’s tool-use policy into four modules, including Planner, Selector, Caller, and Synthesizer, and iteratively improves them in a self-improving loop through three novel mechanisms. \textit{Trajectory-Grounded Blame Attribution} uses diagnostic traces to localize failures to a specific module. \textit{Feedback-Guided Targeted Mutation} then edits only that module via natural-language critique. \textit{Diversity-Aware Population Selection} preserves complementary candidates to ensure solution diversity. Across four benchmarks, \textsc{EvoTool} outperforms strong baselines by over 5 points on both GPT-4.1 and Qwen3-8B, while achieving superior efficiency and transferability. The code will be released once paper is accepted.

\end{abstract}

\section{Introduction}

LLM-based agents augmented with external tools have become a central paradigm for solving complex tasks in reasoning \cite{wei2022chain, zhou2023language}, decision-making \cite{yao2022react, xie2024osworld}, and domain-specific automation \cite{bran2023chemcrow, yang2024swe}. 
These agents depend on an effective tool-use policy to coordinate interdependent competencies, including goal decomposition, tool selection, argument construction, and the grounded synthesis of tool outputs \cite{qu2025tool}.
However, achieving reliable tool use in practice remains challenging, as real-world tasks often involve long-horizon, tightly coupled decision chains where a single error in planning, selection, invocation, or synthesis can cause overall failure \cite{liu2023agentbench}. 
At the same time, supervision is typically available only at the end of an interaction \cite{shinn2023reflexion}, collapsing multiple error sources into a single terminal signal. 
Together, these factors create a severe credit-assignment problem that obscures the specific cause of failure and hinders targeted policy improvement.

\begin{figure}[!t]
    \centering
    \includegraphics[width=1\linewidth]{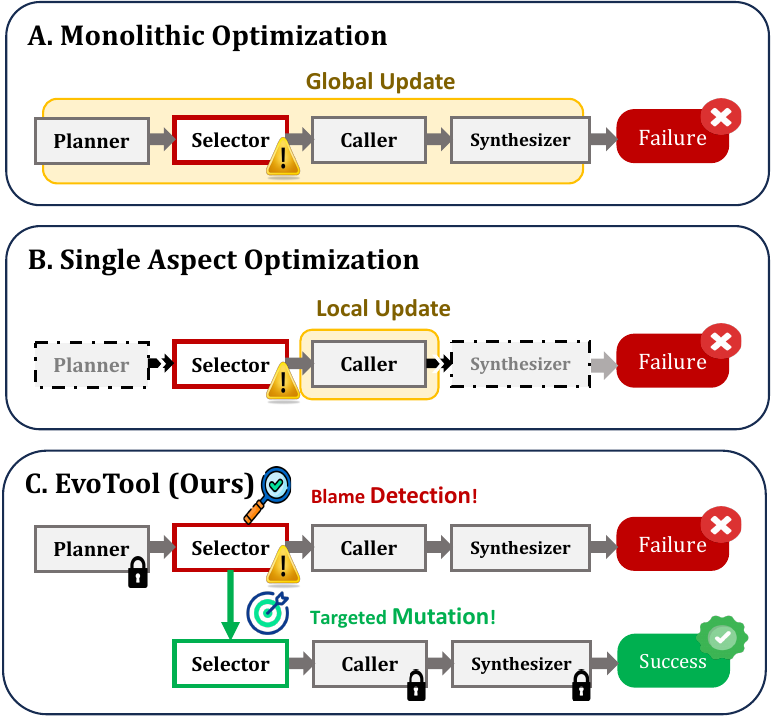}
    \caption{Limitations of monolithic and single-aspect tool-use policy optimization, and how EVOTOOL enables targeted, blame-aware policy updates.}
    \vspace{-2em}
    \label{fig:Intro_Example}
\end{figure}

Early approaches rely on hand-crafted prompting patterns or fixed control heuristics to implement tool-use policy \cite{yao2022react, wang-etal-2023-plan}, which require substantial manual effort and lead to system collapse when unanticipated errors disrupt the rigid execution flow.
More recent work explores automated policy optimization \cite{gao2025survey} but generally diverges into two extremes that fail to resolve the credit assignment dilemma as illustrated in Figure~\ref{fig:Intro_Example}. Monolithic Policy optimization methods \cite{yang2023large,fernando2023promptbreeder} apply global black-box search over the entire agent prompt, which can entangle heterogeneous behaviors across modules and induce regressions where fixing one error destabilizes other capabilities. In contrast, single-aspect optimization methods refine individual components in isolation, such as planning or tool calling \cite{sun2023adaplanner,yuan2025easytool}, while ignoring cross-module error propagation in long-horizon trajectories.
Consequently, no existing paradigm simultaneously achieves the targeted error correction and multi-module coordination needed for robust tool-use policy optimization.

In light of this, we propose \textcolor{black}{\textsc{EvoTool}}, a novel self-evolving framework that iteratively optimizes a modular tool-use policy, comprising Planner, Selector, Caller, and Synthesizer, through a gradient-free evolutionary paradigm. 
\textcolor{black}{Specifically, instead of relying on a single terminal output, we propose \emph{Trajectory-Grounded Blame Attribution}, which exploits intermediate feedback from the tool environment to convert trace-level diagnostics into module-level responsibility signals, identifying the component most likely to have caused the failure. Once the target module is identified, a \emph{Feedback-Guided Targeted Mutation} mechanism then generates a trace-grounded natural-language feedback to selectively edit the blamed module specification, while keeping the remaining modules fixed.}

Furthermore, since tool-use competence decomposes into multiple partially competing skills, greedily selecting a single 'best' candidate by global average can discard complementary behaviors and induce premature convergence. Instead, \textsc{EvoTool} adopts a \textit{Diversity-Aware Population Selection} strategy, which retains a population of policy variants and selects candidates based on instance-level wins, thereby preserving complementary strengths while discouraging collapse to a narrow strategy.
Together, \textsc{EvoTool} resolves key limitations of prior work by avoiding the instability of monolithic global edits and the incompleteness of single-aspect refinement, addressing credit assignment under delayed feedback to enable coordinated, targeted improvement across interdependent competencies. Experiments on four diverse benchmarks show that our approach consistently outperforms SoTA baselines by over 5 points on both open-source and closed-source models while achieving superior token efficiency and transferability across both datasets and models.

Main contributions are shown as follows:
\begin{itemize}
    \item We propose \textbf{\textcolor{black}{\textsc{EvoTool}}}, a self-evolving framework that optimizes a modular tool-use policy using a gradient-free evolutionary paradigm.
    \item We introduce Trajectory-Grounded Blame Attribution and Feedback-Guided Targeted Mutation, which leverage structured trajectories and natural-language critique to produce localized updates to the responsible module.
    \item We propose Diversity-Aware Population Selection to preserve complementary candidates across heterogeneous tool-use competencies.
    \item We evaluate \textsc{EvoTool} on four diverse benchmarks, showing consistent performance gains over state-of-the-art baselines alongside superior token efficiency and robust transferability across models and datasets.
\end{itemize}

\section{Related Work}
\paragraph{Agent Tool-Use Policy Learning.} Agent tool-use policy has progressed from static engineering to dynamic self-improvement \cite{jiang2025adaptation}. Early systems relied on hand-crafted prompting patterns and fixed control heuristics \cite{yao2022react,wei2022chain, wang-etal-2023-plan, shen2023hugginggpt, lu2025octotools}, which demand excessive manual effort and often generalize poorly across domains and tools. Training-based approaches internalize tool use via SFT \cite{schick2023toolformer, patil2024gorilla} or RL \cite{qian2025toolrl, feng2025retool}, but adaptation to evolving environments remains costly due to static weights and large data requirements.
A complementary direction therefore explores training-free optimization, refining tool-use behavior online from interaction feedback without weight updates \cite{ramnath-etal-2025-systematic}. However, these methods face a key trade-off: global edits can entangle heterogeneous behaviors in planning and execution \cite{yang2023large, guo2023connecting}, while isolated local optimizations \cite{qu2024exploration,du2024anytool,yuan2025easytool} ignore cross-module error propagation in long-horizon trajectories. We therefore explore gradient-free search that couples blame attribution with targeted mutation to localize failures and repair the responsible module without destabilizing the overall policy.

\paragraph{Self-Evolving Agent Systems.} Self-evolving agent systems enable agents to acquire and refine competencies over time by repeatedly acting, evaluating outcomes, and updating decision policies \cite{gao2025survey, fang2025comprehensive}. This line of work includes self-reflection and self-correction \cite{madaan2023self, shinn2023reflexion} that turn failures into natural-language or structured feedback \cite{yuksekgonul2024textgrad,agrawal2025gepa}, as well as continual improvement loops that accumulate skills or update agent components \cite{khattab2023dspy,hu2024automated}. However, many frameworks rely on greedy selection and converge prematurely to a narrow strategy, discarding diverse behaviors needed for heterogeneous task distributions \cite{fernando2023promptbreeder}. Therefore, we explore diversity-aware population selection to maintain a heterogeneous candidate pool.

\section{Preliminaries}

\paragraph{Task and Environment Formulation.}
We study LLM-based agents that solve tasks by interacting with an external tool environment. Let $x \sim \mathcal{D}$ denote a task instance drawn from a distribution $\mathcal{D}$, and let $\mathcal{T}=\{\tau_1,\dots,\tau_K\}$ be a set of tools, each specified by a name, an argument schema, and a documentation. At step $t$, the agent observes a textual state $s_t$ summarizing the interaction history and outputs an action $a_t$, which either performs intermediate reasoning, invokes a tool or terminates with an answer. The environment $E$ then executes tool invocations and returns an observation $o_t$ with tool outputs. A complete agent run yields a trajectory $\tau=\{(s_t,a_t,o_t)\}_{t=1}^{T}$ of variable length $T$, which terminates when the agent outputs a final prediction $\hat{y}=\hat{y}(\tau)$. We evaluate performance using a task-dependent success function $R(x,\hat{y}) \in [0,1]$ that is instantiated by standard metrics such as pass@1, success rate, and F1 score. For convenience, we summarize each rollout as an episode record $e=(x,\tau,\hat{y}, R(x,\hat{y}))$. 


\paragraph{Modularized Tool-Use Policy.}
We represent tool use as a modular policy with four roles: (i) a planner that decomposes the input into subgoals; (ii) a selector that decides whether and which tool to call given the current state $s_t$ and subgoal; (iii) a caller that constructs valid arguments and executes the selected tool; and (iv) a synthesizer that integrates tool outputs into the final response. Accordingly, the overall tool-use policy is defined as a fixed modular composition:
$
\Pi \;=\; \pi_{\text{syn}} \circ \big(\pi_{\text{call}} \circ (\pi_{\text{sel}} \circ \pi_{\text{plan}})\big),
$ where each module operates on the intermediate state produced by the previous one.
All modules share the same base LLM with frozen weights $W$, conditioned on evolvable module specifications $\Theta=\{\theta_{\text{plan}},\theta_{\text{sel}},\theta_{\text{call}},\theta_{\text{syn}}\}$ including prompts, tool templates, or lightweight formatting rules. We denote the instantiated policy by $\pi_{\Theta,W}$. Learning updates only $\Theta$; the model weights $W$ remain fixed.

\paragraph{Optimization Objective.}
Executing $\pi_{\Theta,W}$ in environment on input $x$ induces a trajectory $\tau \sim (\pi_{\Theta,W},E)\mid x$ and a terminal answer $\hat{y}(\tau)$. Our objective is to maximize expected task success by evolving $\Theta$ under frozen weights $W$:
\begin{equation}
J(\Theta;W) \;=\; \mathbb{E}_{x \sim \mathcal{D}} \; \mathbb{E}_{\tau \sim (\pi_{\Theta,W})\mid x}
\big[\, R\big(x,\hat{y}(\tau)\big) \,\big].
\end{equation}
Therefore, the core challenge is: \textit{how can we improve the tool-use policy from sparse, end-of-trajectory outcomes while (i) localizing which component caused the failure and (ii) updating only that component to avoid breaking other behaviors?}

\begin{figure*}
    \centering
    \includegraphics[width=\linewidth]{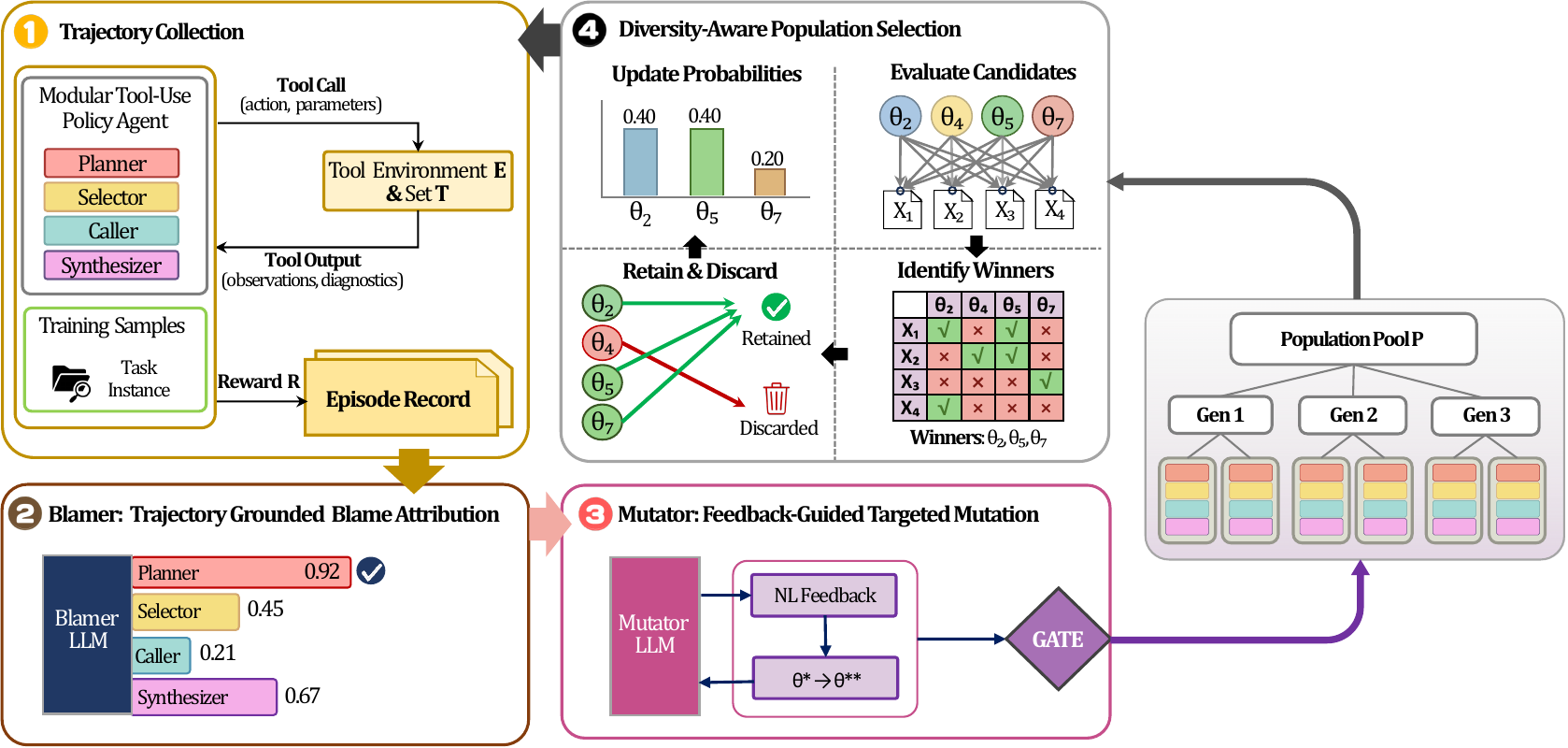}
    \caption{Overall architecture of \textsc{EvoTool}. \textsc{EvoTool} optimizes a modular tool-use policy through a self-evolving loop consisting of (1) trajectory collection from the tool environment, (2) trajectory-grounded blame attribution to identify the responsible module, (3) feedback-guided targeted mutation to update that module using natural language feedback, and (4) diversity-aware population selection over candidate policies to retain complementary candidates.}
    \label{fig:architecture}
\end{figure*}

\section{Methods}

In this section, we introduce \textcolor{black}{\textsc{EvoTool}}, a self-evolving framework designed to optimize the modular tool-use policy through a gradient-free evolutionary paradigm. \textcolor{black}{\textsc{EvoTool}} maintains a population of candidate module specifications and iteratively improves them using three mechanisms: (i) Trajectory-Grounded Blame Attribution to localize specific responsible module, (ii) Feedback-Guided Targeted Mutation to update only the blamed module using dense feedback signal, and (iii) Diversity-Aware Population Selection to preserve candidates with complementary competences.

\paragraph{Self-Evolving Optimization Loop}

As shown in Figure \ref{fig:architecture}, \textcolor{black}{\textsc{EvoTool}} optimizes tool-use policy by iteratively refining module specifications $\Theta=\{\theta_{\text{plan}},\theta_{\text{sel}},\theta_{\text{call}},\theta_{\text{syn}}\}$ while keeping base LLM weights frozen. We maintain a population $\mathcal{P}=\{\Theta^{(i)}\}_{i=1}^{N}$ of candidate specifications. In each generation, we sample a parent $\Theta\in\mathcal{P}$ and execute the instantiated tool-use policy $\pi_{\Theta,W}$ on a mini-batch of task instances drawn from training pool $S_{\text{train}}$, collecting the corresponding episode records $e=\big(x,\tau,\hat{y},R(x,\hat{y})\big)$. To translate these interaction traces into policy updates, we first identify the target module $\pi^{\ast}\in\Pi$ using a trajectory-grounded blame score $b_{\pi}(e)$. We then construct dense, natural-language feedback $F(e,\pi^{\ast})$ to edit only the corresponding module specification, producing a child candidate $\Theta'$ that differs from its parent in exactly one component. The child $\Theta'$ is added to $\mathcal{P}$ only if it outperforms the parent on the mini-batch. Finally, we update the parent sampling distribution using diversity-aware selection based on evaluation on a held-out set for population selection $S_{\text{sel}}$. We repeat this loop for a fixed budget and return the best-performing candidate in $\mathcal{P}$. Algorithm details are in Appendix \ref{app:algo}.

\subsection{Trajectory-Grounded Blame Attribution}
\label{sec: blame attribution}

Failures in long-horizon, multi-tool settings are often heterogeneous and multi-causal: a single negative outcome can originate from diverse sources, such as poor goal decomposition, incorrect tool selection, schema violations, or ungrounded synthesis. Without an explicit diagnostic mechanism, a self-evolving agent cannot reliably determine which module to update, degenerating optimization into
global blind search. Therefore, \textcolor{black}{\textsc{EvoTool}} converts each episode record $e$ into a localized repair target by assigning module-level blame to isolate a single component for mutation. Prompt details on Blamer LLM can be found in Appendix \ref{app: Blamer Prompt}.


Given an episode record $e=(x,\tau,\hat{y},R(x,\hat{y}))$, we first extract structured diagnostic events from the trajectory $\tau=\{(s_t,a_t,o_t)\}_{t=1}^{T}$, including tool-choice outcomes, argument validity signals, tool execution outcomes, and synthesis-grounding signals. We next provide the full episode $e$ together with this diagnostic summary to a \textit{Blamer} LLM, which outputs module-wise blame scores $b_{\pi}(e) \in [0,1]$, 
 where large $b_{\pi}(e)$ indicates stronger evidence that module policy $\pi$ is responsible for the observed failure or suboptimality. We then select the module $\pi^{\ast}$ with the highest blame score as the mutation target and pass it, along with the episode record $(e,\pi^{\ast})$, for a later targeted mutation. 


\subsection{Feedback-Guided Targeted Mutation}
\label{sec: targed mutation}

Given a blamed episode $e$ and its selected target module $\pi^{\ast}$, \textcolor{black}{\textsc{EvoTool}} produces a child policy by editing only the corresponding module specification while freezing the remaining components. 
This targeted design directly mitigates the sparsity and delay of $R(x,\hat{y})$ by leveraging the full interaction trace as supervision: we prompt a reflective \textit{mutator} LLM to translate the blamed episode into natural-language feedback $F(e,\pi^{\ast})$ that both explains the error mode from the perspective of the selected module and proposes a concrete, localized edit to that module's specification given the current specification and trajectory evidence. We then apply this edit to form a child candidate $\Theta'$, keeping all other modules fixed.
By restricting updates to a single blamed module and using rich textual feedback grounded in trajectories, \textcolor{black}{\textsc{EvoTool}} minimizes unintended regressions in unrelated competencies while enabling interpretable, trace-grounded improvements. Prompt details for Mutator LLM can be found in Appendix \ref{app: Mutator Prompt}.



\begin{table*}[!t]
\centering
\footnotesize
\renewcommand{\arraystretch}{1.05}

\resizebox{\textwidth}{!}{%
\begin{tabular}{c|l|cccc|ccc|ccc|ccc|c}
\toprule
\multicolumn{2}{c|}{\textbf{Model / Method}} &
\multicolumn{4}{c|}{\textbf{ToolBench}} &
\multicolumn{3}{c|}{\textbf{RestBench}} &
\multicolumn{3}{c|}{\textbf{$\tau$-Bench}} &
\multicolumn{3}{c|}{\textbf{BFCL}} &
\textbf{Overall} \\
\multicolumn{2}{c|}{} &
G1 & G2 & G3 & Avg &
\textcolor{black}{TM} & \textcolor{black}{SP} & Avg &
\textcolor{black}{Ret} & \textcolor{black}{Air} & Avg &
\textcolor{black}{Sin} & \textcolor{black}{Mul} & Avg &
Avg \\
\midrule

\multirow{15}{*}{\rotatebox{90}{\textbf{GPT-4.1}}}
& \multicolumn{1}{l}{\cellcolor{gray!15}\textbf{Hand-crafted policies}}
& \multicolumn{14}{c}{\cellcolor{gray!15}} \\
& \quad ReAct         & 68.2 & 64.7 & 57.9 & 63.6 & 72.4 & 74.3 & 73.4 & 59.8 & 35.9 & 47.9 & 74.8 & 37.2 & 56.0 & 60.6 \\
& \quad CoT           & 53.5 & 50.8 & 48.4 & 50.9 & 59.8 & 63.5 & 61.7 & 34.2 & 25.3 & 29.8 & 43.2 & 21.4 & 32.3 & 44.5 \\
& \quad Plan-and-Solve  & 63.3 & 59.2 & 55.3 & 59.3 & 64.8 & 69.2 & 67.0 & 58.4 & 36.7 & 47.6 & 57.2 & 25.3 & 41.3 & 54.4 \\

& \multicolumn{1}{l}{\cellcolor{gray!15}\textbf{Monolithic optimization}}
& \multicolumn{14}{c}{\cellcolor{gray!15}} \\
& \quad OPRO          & 69.3 & 67.3 & 58.9 & 65.2 & 73.8 & 76.4 & 75.1 & 58.2 & 36.8 & 47.5 & 82.5 & 35.3 & 58.9 & 62.1 \\
& \quad PromptBreeder & 68.5 & 66.5 & 54.7 & 63.2 & 74.1 & 75.2 & 74.7 & 56.4 & 31.3 & 43.9 & 81.3 & 36.2 & 58.8 & 60.5 \\
& \quad EvoPrompt     & 70.1 & 69.8 & 59.2 & 66.4 & 76.3 & 77.4 & 76.9 & 60.8 & 36.3 & 48.6 & \textbf{84.2} & 39.9 & 62.1 & 63.8 \\

& \multicolumn{1}{l}{\cellcolor{gray!15}\textbf{\textcolor{black}{Single-aspect optimization}}}
& \multicolumn{14}{c}{\cellcolor{gray!15}} \\
& \quad AdaPlanner    & 62.2 & 58.5 & 48.8 & 56.5 & 66.2 & 70.1 & 68.2 & 62.2 & 38.8 & 50.5 & 69.2 & 41.1 & 55.2 & 57.5 \\
& \quad \textsc{EasyTool}      & 80.3 & 75.1 & 66.2 & 73.9 & 81.4 & 83.5 & 82.5 & 51.2 & 29.9 & 40.6 & 76.1 & 36.0 & 56.1 & 64.4 \\
& \quad DRAFT         & 82.1 & 76.8 & 68.4 & 75.8 & 84.3 & 85.2 & 84.8 & 48.4 & 29.2 & 38.8 & 77.8 & 32.0 & 54.9 & 64.9 \\
& \quad \textsc{AnyTool}       & 73.3 & 68.8 & 61.0 & 67.7 & 75.2 & 78.4 & 76.8 & 60.4 & 36.2 & 48.3 & 77.1 & 39.3 & 58.2 & 63.3 \\

& \multicolumn{1}{l}{\cellcolor{gray!15}\textbf{Ours}}
& \multicolumn{14}{c}{\cellcolor{gray!15}} \\
& \quad \textbf{\textcolor{black}{\textsc{EvoTool}}} & \textbf{83.5} & \textbf{78.2} & \textbf{71.5} & \textbf{77.7} & \textbf{86.2} & \textbf{86.1} & \textbf{86.2} & \textbf{64.8} & \textbf{39.1} & \textbf{52.0} & \underline{83.9} & \textbf{42.3} & \textbf{63.1} & \textbf{70.6} \\
\midrule

\multirow{15}{*}{\rotatebox{90}{\textbf{Qwen3-8B}}}
& \multicolumn{1}{l}{\cellcolor{gray!15}\textbf{Hand-crafted policies}}
& \multicolumn{14}{c}{\cellcolor{gray!15}} \\
& \quad ReAct         & 60.0 & 55.0 & 47.5 & 54.2 & 63.7 & 63.2 & 63.5 & 33.1 & 14.4 & 23.8 & 70.6 & 33.3 & 52.0 & 49.0 \\
& \quad CoT           & 47.1 & 43.2 & 39.7 & 43.3 & 52.6 & 54.0 & 53.3 & 14.6 & 8.2 & 11.4 & 41.9 & 20.3 & 31.1 & 35.7 \\
& \quad Plan-and-Solve  & 55.7 & 50.3 & 45.3 & 50.4 & 57.0 & 58.8 & 57.9 & 32.0 & 15.0 & 23.5 & 55.5 & 24.0 & 39.8 & 43.7 \\

& \multicolumn{1}{l}{\cellcolor{gray!15}\textbf{Monolithic optimization}}
& \multicolumn{14}{c}{\cellcolor{gray!15}} \\
& \quad OPRO          & 61.0 & 57.2 & 48.3 & 55.5 & 64.9 & 64.9 & 64.9 & 31.9 & 15.0 & 23.5 & 75.2 & 33.5 & 54.4 & 50.2 \\
& \quad PromptBreeder & 60.3 & 56.5 & 44.9 & 53.9 & 65.2 & 63.9 & 64.6 & 30.6 & 11.3 & 21.0 & 73.9 & 34.4 & 54.2 & 49.0 \\
& \quad EvoPrompt     & 61.7 & 59.3 & 48.5 & 56.5 & 67.1 & 65.8 & 66.5 & 33.8 & 14.7 & 24.3 & 75.7 & 35.9 & 55.8 & 51.4 \\

& \multicolumn{1}{l}{\cellcolor{gray!15}\textbf{\textcolor{black}{Single-aspect optimization}}}
& \multicolumn{14}{c}{\cellcolor{gray!15}} \\
& \quad AdaPlanner    & 54.7 & 49.7 & 40.0 & 48.1 & 58.3 & 59.6 & 59.0 & 34.8 & \textbf{16.4} & 25.6 & 67.1 & 36.0 & 51.6 & 46.3 \\
& \quad \textsc{EasyTool}      & 70.7 & 61.8 & 54.3 & 62.3 & 71.6 & 68.0 & 69.8 & 20.9 & 10.3 & 15.6 & 73.8 & 28.2 & 51.0 & 51.1 \\
& \quad DRAFT         & 72.2 & 65.3 & 56.1 & 64.5 & 74.2 & 72.4 & 73.3 & 17.8 & 8.9 & 13.4 & 72.5 & 27.1 & 49.8 & 51.8 \\
& \quad \textsc{AnyTool}       & 64.5 & 58.5 & 50.0 & 57.7 & 66.2 & 66.6 & 66.4 & 23.5 & 14.6 & 19.1 & 71.8 & 30.3 & 51.1 & 49.6 \\

& \multicolumn{1}{l}{\cellcolor{gray!15}\textbf{Ours}}
& \multicolumn{14}{c}{\cellcolor{gray!15}} \\
& \quad \textbf{\textcolor{black}{\textsc{EvoTool}}} & \textbf{73.5} & \textbf{66.5} & \textbf{58.6} & \textbf{66.2} & \textbf{75.9} & \textbf{73.2} & \textbf{74.6} & \textbf{35.9} & \underline{15.7} & \textbf{25.8} & \textbf{76.9} & \textbf{36.4} & \textbf{56.7} & \textbf{57.0} \\
\bottomrule
\end{tabular}%
}

\caption{Main results on two backbone models. We report benchmark-standard metrics on ToolBench (G1/G2/G3), RestBench (TMDB/Spotify), $\tau$-Bench (Retail/Airline), and BFCL (Single/Multi-turn), with the overall average. TM = TMDB, SP = Spotify, Ret = Retail, Air = Airline, Sin = Single-turn, Mul = Multi-turn.}
\label{tab:main_results_streams}
\end{table*}

\textcolor{black}{\subsection{Diversity-Aware Population Selection}}
\label{sec: population selection}



To prevent the population from collapsing into a single mode and forgetting previously mastered behaviors, \textcolor{black}{\textsc{EvoTool}} explicitly preserves candidates with complementary competencies. Instead of greedily selecting parents based on global average performance, we employ an instance-wise winner criterion on a held-out set $S_{\text{sel}}$. After each generation, we evaluate all candidates $\Theta^{(i)} \in \mathcal{P}$ on $S_{\text{sel}}$. A candidate is retained only if it achieves the highest score $R(x, \hat{y})$ on at least one instance $x \in S_{\text{sel}}$. Candidates that never achieve instance-level dominance are removed, as they do not represent distinct competencies under the evaluation distribution. For the remaining population, we sample parents for subsequent generations proportional to their winner frequency (the fraction of instances where they outperform all others), thereby concentrating updates on broadly effective policies while retaining specialists that cover distinct regions of the task distribution. Implementation details can be found in Appendix \ref{app: implementation details}.

\begin{figure*}[!t]
    \centering

    \begin{subfigure}{\textwidth}
        \centering
        \includegraphics[width=1\linewidth]{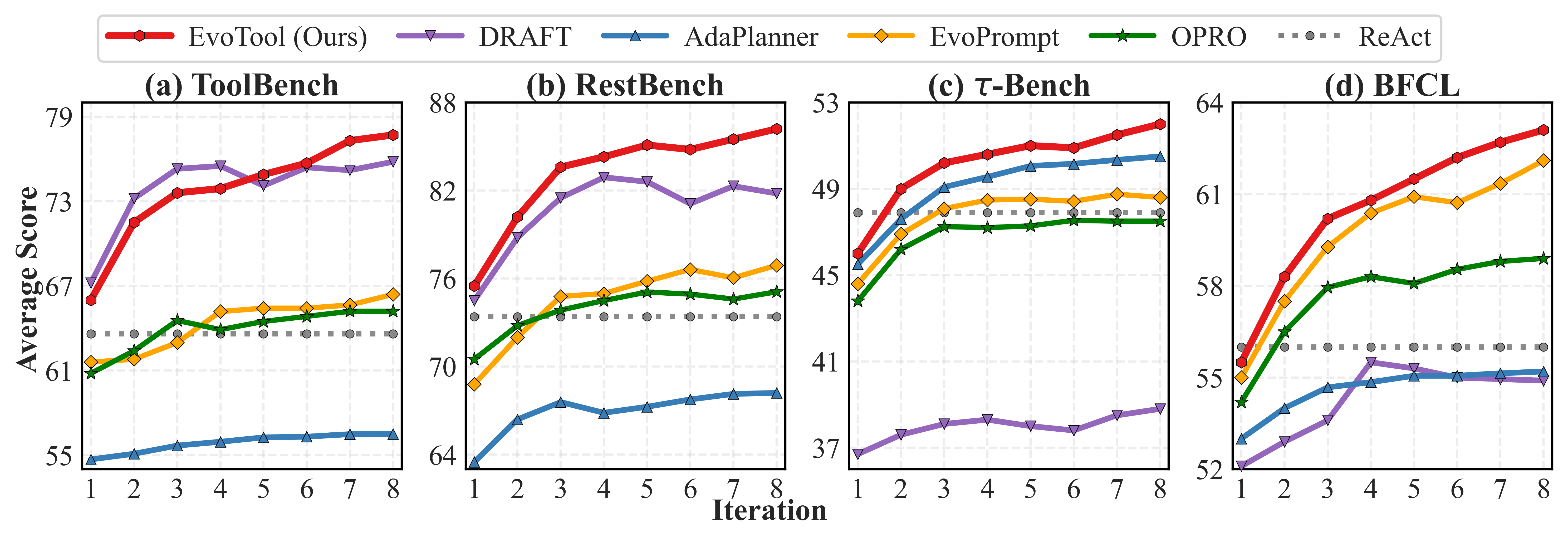}
        \label{fig:learning_curve}
    \end{subfigure}
    \vspace{-12mm} 

    \begin{subfigure}{\textwidth}
        \centering
        \includegraphics[width=1\linewidth]{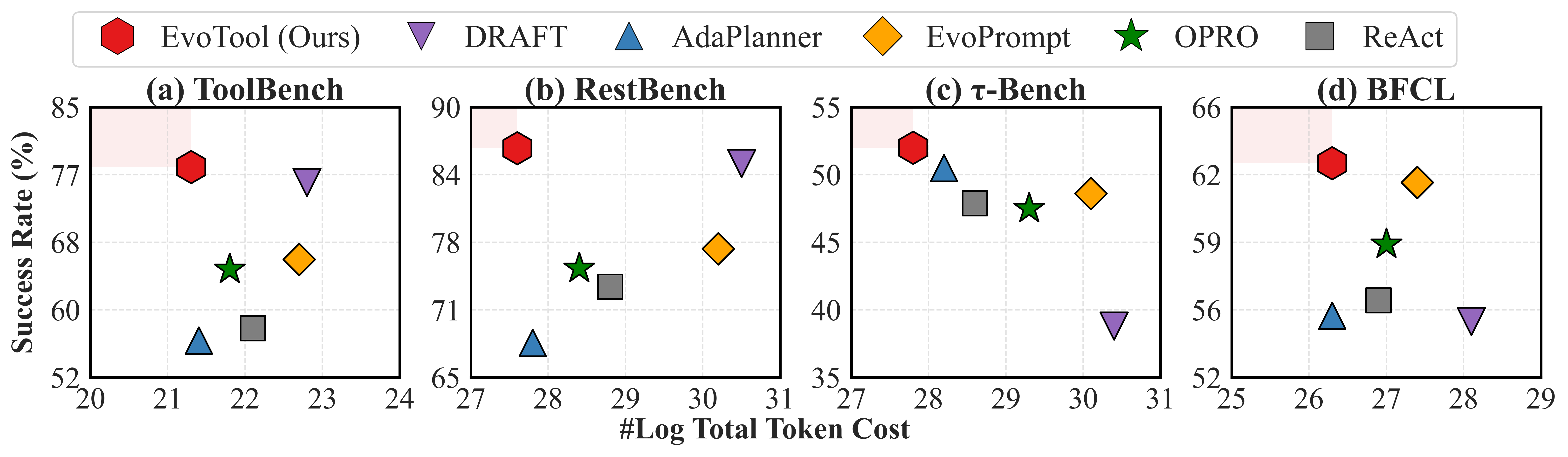}
        \label{fig:Efficiency}
    \end{subfigure}

    \vspace{-3mm} 
    \caption{\textbf{Learning dynamics and efficiency comparison.} (a) Learning curves across evolution iterations on four benchmarks. (b) Performance versus log token cost under GPT-4.1.}
    \label{fig:lc_efficiency}
\end{figure*}

\section{Experiment}
\subsection{Experiment Setup}

\paragraph{Benchmarks and Metrics.} 
We evaluate \textsc{EvoTool} on four established tool-use benchmarks following their standard setups. On \textbf{ToolBench} \cite{qin2023toolllm}, which tests large-scale API generalization over RapidAPI, we report \emph{Pass Rate} on the G1/G2/G3 subsets. On \textbf{RestBench} \cite{song2023restgpt}, which evaluates sequential tool use over real REST APIs, we report \emph{Success Rate} on the TMDB and Spotify subsets. On \textbf{\(\bm{\tau}\)-Bench} \cite{yao2024tau}, which evaluates stateful, long-horizon agent--user interactions in the retail and airline domains, we report \emph{Pass@1} to assess single-rollout success. On \textbf{BFCL} \cite{patil2025bfcl}, which assesses function-calling for tool invocation, we report \emph{Accuracy} on the selected single-turn and multi-turn subsets. For each benchmark, we additionally report the within-benchmark average over its subsets and an overall average across benchmarks. Details on the selected benchmarks can be found in the Appendix \ref{app: datasets}.

\paragraph{Baselines.}
We compare \textsc{EvoTool} against three streams of tool-use policy design.
\textbf{(1) Hand-crafted tool-use policy} that relies on manually designed prompting patterns and fixed control heuristics, including \textsc{ReAct} \cite{yao2022react}, chain-of-thought prompting \cite[CoT,][]{wei2022chain}, and \textsc{Plan-and-Solve} \cite{wang-etal-2023-plan} pipelines.
\textbf{(2) Monolithic tool-use policy optimization} that treats agent prompt as a single global policy object optimized via black-box prompt search, including \textsc{OPRO} \cite{yang2023large}, \textsc{PromptBreeder} \cite{fernando2023promptbreeder}, and \textsc{EvoPrompt} \cite{guo2023connecting}.
\textbf{(3) \textcolor{black}{Single-aspect (local/partial) tool-use policy optimization}} that optimize individual component of tool-use policy in isolation, including \textsc{AdaPlanner} \cite{sun2023adaplanner} that refines tool planning independently, \textsc{DRAFT} \cite{qu2024exploration} and \textsc{EasyTool} \cite{yuan2025easytool} that improves tool selection and calling separately and \textsc{AnyTool} \cite{du2024anytool} that refines selection and answer synthesizing.
Across all baselines, we use the same base LLM, tool suite, and evaluation budget for a fair comparison. Across each baseline, we select GPT-4.1 \cite{openai2025gpt41} and Qwen3-8B \cite{yang2025qwen3} as the backbone model to verify its generalizability.
Further details on the selected baselines can be found in Appendix \ref{app:baselines}.

\subsection{Results}

As shown in Table~\ref{tab:main_results_streams}, \textsc{EvoTool} consistently outperforms all baselines across both base models and four benchmarks, demonstrating superior policy learning. On the stronger GPT-4.1 backend, our framework achieves an overall average of 70.6, surpassing the strongest single-aspect baseline DRAFT by nearly 6 points and the best monolithic baseline EvoPrompt by roughly 7 points; this dominance extends to Qwen3-8B, where it outperforms the second best baseline by 5.2 points. Crucially, \textsc{EvoTool} remains robust on complex, long-horizon tasks where other paradigms falter: on the stateful $\tau$-Bench, single-aspect methods like DRAFT and \textsc{EasyTool} drop to 38.8 and 40.6 as isolated optimization misses cross-module dependencies, while \textsc{EvoTool} reaches 52.0. This advantage is reinforced by a leading 42.3 on BFCL multi-turn, confirming that \textsc{EvoTool} balances planning flexibility and syntactic precision for deep reasoning better than other alternatives.


\subsection{Learning Dynamics and Efficiency}


Figure \ref{fig:lc_efficiency} presents a comprehensive view of the learning dynamics and efficiency of \textsc{EvoTool} compared to representative baselines. In terms of performance progression (Figure \ref{fig:lc_efficiency}a), \textsc{EvoTool} delivers the most consistent gains, overtaking baselines and sustaining a monotonic upward trajectory to the highest final scores. This contrasts sharply with single-aspect methods like DRAFT and AdaPlanner, which exhibits clear domain brittleness; although they excel in specific niches such as simple APIs or planning-intensive tasks, they stagnate elsewhere. Similarly, monolithic optimizers like OPRO and EvoPrompt struggle with broad API generalization due to interference from global updates. Simultaneously, in terms of cost-effectiveness (Figure \ref{fig:lc_efficiency}b), \textsc{EvoTool} consistently dominates the optimal upper-left quadrant, delivering superior performance with minimal token usage. While baselines like EvoPrompt and OPRO require substantially higher token costs to reach competitive scores, \textsc{EvoTool} avoids such overhead by restricting mutations to targeted components, effectively decoupling capability growth from token inflation.

\section{Ablation}

\begin{table}[t]
\centering
\small
\setlength{\tabcolsep}{2.5pt}
\begin{adjustbox}{max width=\columnwidth}
\begin{tabular}{l c c c c |c}
\toprule
\textbf{Variant} & \textbf{ToolBench} & \textbf{RestBench} & \boldmath$\bm{\tau}$\textbf{-Bench} & \textbf{BFCL} & \textbf{Avg} \\
\midrule
Static & 55.9 & 65.5 & 21.0 & 52.0 & 48.6 \\
Random & 45.8 & 52.7 & 15.9 & 43.6 & 39.5 \\
\midrule
\textit{Single-Module} \\
\quad Plan-only & 51.2 & 58.6 & 24.7 & 50.1 & 46.2 \\
\quad Sel-only  & 63.1 & 70.8 & 20.4 & 48.2 & 50.6 \\
\quad Call-only & 57.4 & 66.3 & 20.2 & 55.6 & 49.9 \\
\quad Syn-only  & 55.0 & 65.1 & 21.2 & 50.7 & 48.0 \\
\midrule
Monolithic & 59.6 & 67.2 & 20.6 & 48.8 & 49.1 \\
\midrule
\textbf{\textcolor{black}{\textsc{EvoTool}}} & 66.2 & 74.6 & 25.8 & 56.7 & 57.0 \\
\bottomrule
\end{tabular}
\end{adjustbox}
\vspace{-0.3em}
\caption{Blame-targeting ablations (Qwen3-8B). Each variant differs only in how the mutation target is chosen.}
\vspace{-0.5em}
\label{tab:blame_ablation}
\end{table}

\paragraph{Effectiveness of Blame Attribution.} Table~\ref{tab:blame_ablation} evaluates our trajectory-grounded blame attribution on Qwen3-8B. We compare our full framework against (1) \textit{Static} baseline with no evolution, (2) \textit{Random} setting that arbitrarily mutates the target, (3) \textit{Single-Module} variants that always evolve specific components, and (4) \textit{Monolithic} variant that optimizes all modules simultaneously. The results indicate that random mutation brings destructive noise, reducing the average by over 9 points relative to the static baseline. Although single-module variants can improve individual benchmarks, for instance, the Selector-only agent on ToolBench, they generalize poorly and lag on harder settings such as $\bm{\tau}$-Bench. Similarly, monolithic optimization yields only modest gains, suggesting that coarse updates hinder effective module correction. In contrast, \textsc{EvoTool} achieves the best overall average, showing that blame-based targeting is critical for robust multi-task tool-policy learning.

\paragraph{Effectiveness of Feedback-Guided Mutation.} Table~\ref{tab:ablate_feedback} ablates the two mutation guidance signals: trajectory evidence $\tau$ and explicit natural-language feedback $F$. Removing both yields the lowest success rate, showing that unguided mutations are ineffective for refining complex policies. Dropping feedback causes a larger degradation than dropping trajectory evidence, indicating that explicit critique provides a more informative optimization signal than raw trajectory context. However, the full model performs best, confirming that $\tau$ and $F$ are complementary: as the trajectory trace provides the necessary grounding, while the explicit feedback articulates the optimization direction.

\begin{figure}[!t]
    \centering
    \includegraphics[width=\linewidth]{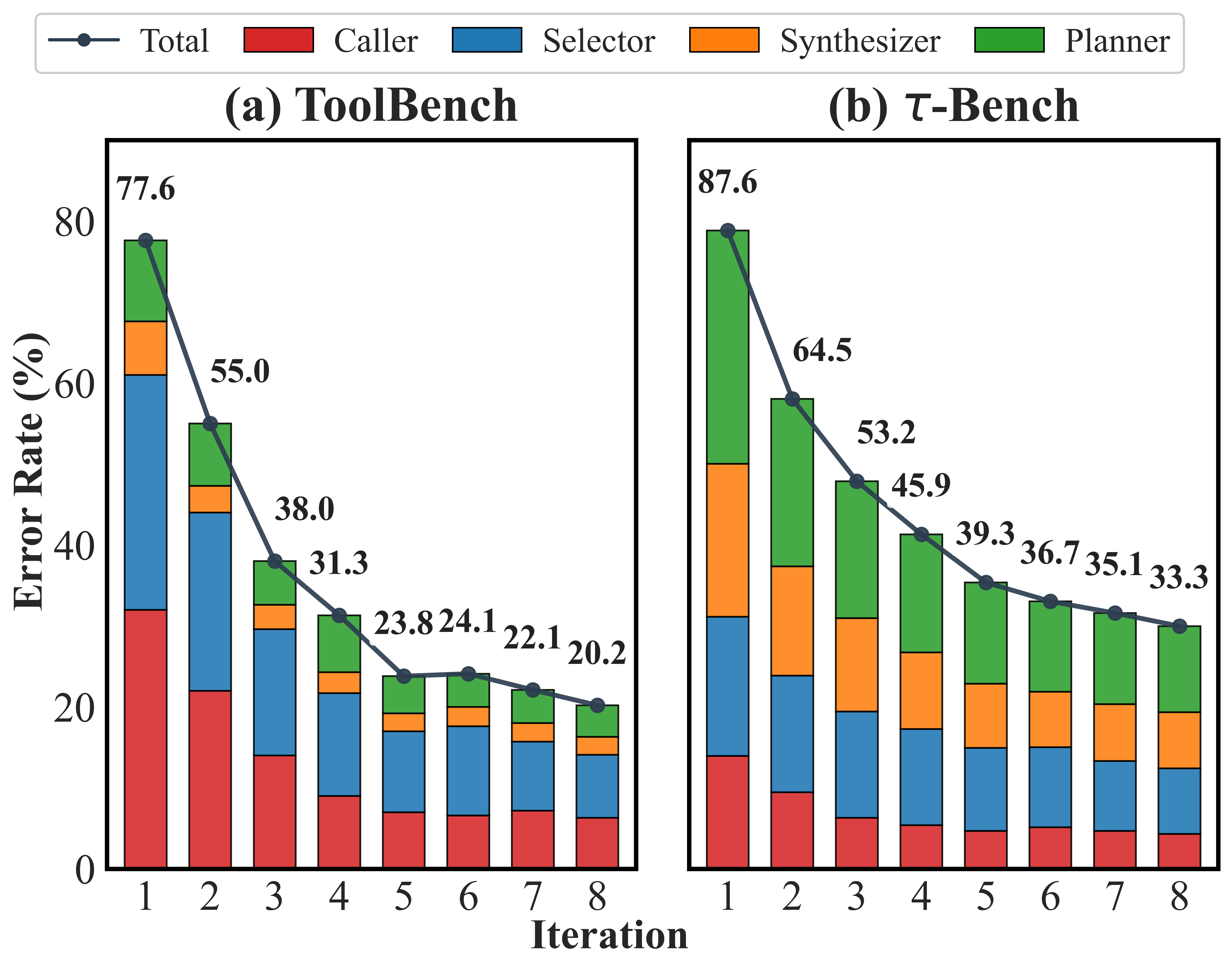}
    \caption{Module-level error progression across evolution iterations diagnosed by the Blamer LLM.}
    \label{fig:error progression}
\end{figure}

\begin{table}[!t]
\centering
\small
\setlength{\tabcolsep}{6pt}
\begin{adjustbox}{width=\columnwidth}
\begin{tabular}{l c c c c}
\toprule
Variant & Uses $\tau$ & Uses $F$ & Success Rate & $\Delta$ vs Full \\
\midrule
\textcolor{black}{\textsc{EvoTool} (full)} & \checkmark & \checkmark & 74.6 & ---- \\
\quad w/o $\tau$ &  & \checkmark & 70.2 & -4.4  \\
\quad w/o $F$& \checkmark &  & 65.2 & -9.4   \\
\quad w/o $\tau$ and $F$ &   &  & 62.8 & -11.8 \\
\bottomrule
\end{tabular}
\end{adjustbox}
\caption{Ablation on mutation guidance. $\tau$ denotes trajectory evidence provided to the mutator; $F$ denotes an explicit natural-language feedback to the mutator.}
\label{tab:ablate_feedback}
\end{table}

\paragraph{Effectiveness of Population Diversity Preservation.} Table~\ref{tab:ablation_selection} examines how parent selection affects evolutionary optimization under identical budgets and mutation settings. Compared to the static baseline, both learning-based selection improves performance, confirming that dynamic evolution is essential for policy learning.
\textsc{EvoTool} performs best across all benchmarks, suggesting its diversity-aware rule better balances exploration and exploitation by retaining high-quality parents while simultaneously preserving sufficient diversity. The advantage is most evident on the harder long-horizon suites on {$\bm{\tau}$-Bench and BFCL, where preserving complementary variants is particularly important.


\section{Analysis}


\paragraph{Error Progression Diagnosis.} Figure~\ref{fig:error progression} decomposes \textsc{EvoTool}’s \emph{Blamer} llm diagnosed errors by module over evolution iterations, showing how targeted, blame-guided mutation prioritizes the most readily correctable failures. On ToolBench, the total error rate drops sharply from 77.6\% to 38.0\% over three iterations, driven mainly by reductions in Caller and Selector errors, aligning with ToolBench’s emphasis on correct tool choice and schema/argument conformity across diverse APIs. As these surface-level issues are resolved, the residual errors concentrate in synthesis and planning, yielding a slower decline to 20.2\% by iteration 8. On $\tau$-Bench, the initial error is higher and remains dominated by Planner-related failures throughout, consistent with its long-horizon, stateful interaction demands. Even after the total error falls to 33.3\%, the persistent planning-heavy profile indicates that multi-step dependency tracking and state maintenance remain the key bottleneck.

\begin{figure}[!t]
    \centering
    \includegraphics[width=1\linewidth]{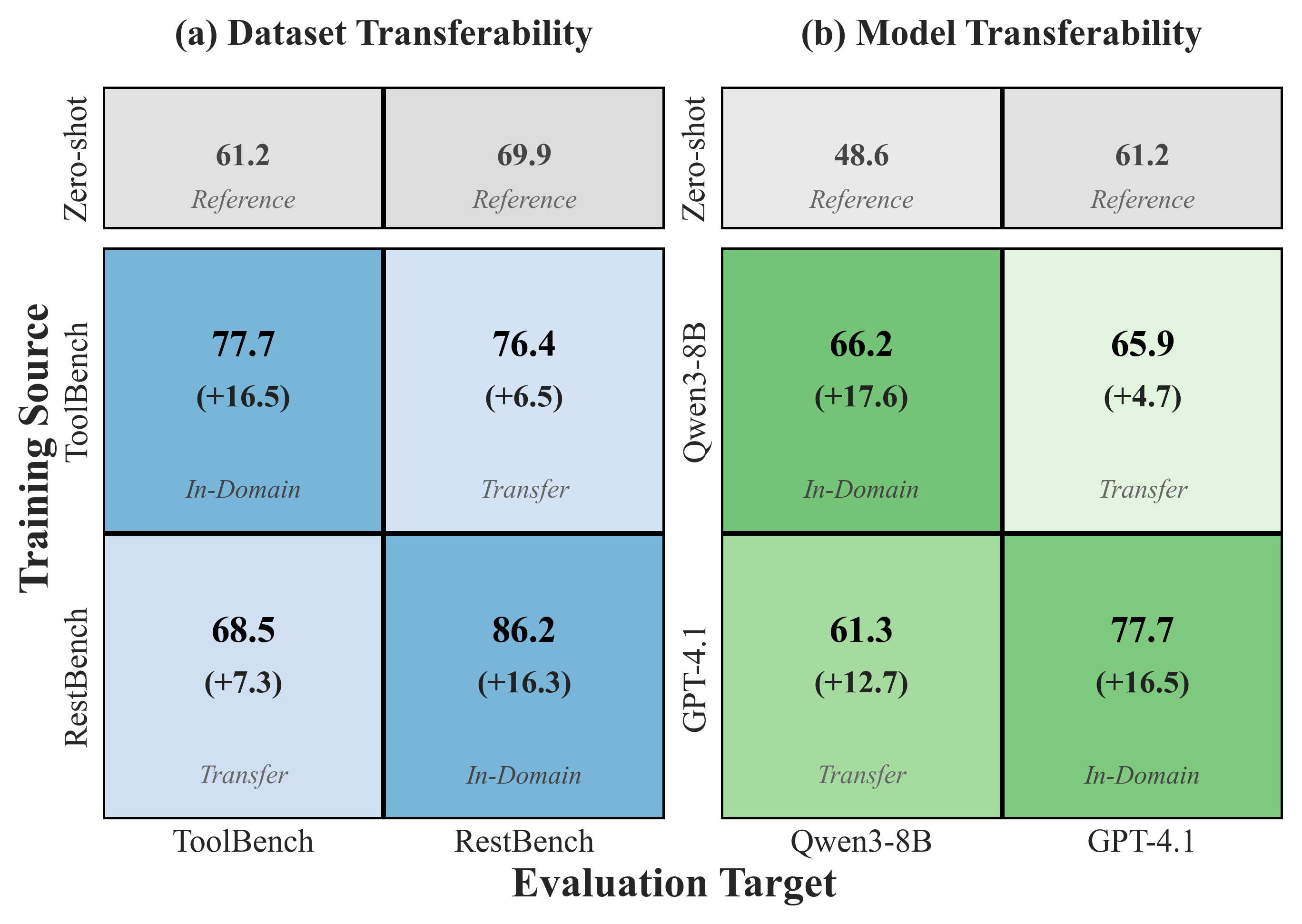}
    \caption{Transferability across datasets and backbone models. (a) Cross-dataset transfer between ToolBench and RestBench. (b) Cross-model transfer between Qwen3-8B and GPT-4.1.}
    \label{fig:transferbility}
\end{figure}

\paragraph{Transferability Across Datasets and Models.} Figure~\ref{fig:transferbility} tests whether \textsc{EvoTool}’s learned policy improvements transfer beyond the setting where they are evolved, using the default static baseline as a reference.
For cross-dataset transfer under GPT-4.1, a policy evolved on ToolBench not only improves in-domain performance but also transfers strongly to RestBench. Similar pattern is observed when RestBench serves as the training set and evaluate on ToolBench.
Besides, cross-model transfer is also robust: a Qwen3-8B-evolved policy remains effective on GPT-4.1, beating baseline by 4.7 points, whereas a GPT-4.1-evolved policy transfers back to Qwen3-8B by a significant 12.7 points. 
Overall, these results suggest \textsc{EvoTool} learns transferable tool-use behaviors that are not narrowly coupled to a single benchmark or model.


\begin{table}[!t]
\centering
\small
\setlength{\tabcolsep}{4.0pt}
\begin{adjustbox}{width=\columnwidth}
\begin{tabular}{l c c c c |c}
\toprule
\textbf{Variant} &
\textbf{ToolBench} &
\textbf{RestBench} &
\textbf{$\bm{\tau}$-Bench} &
\textbf{BFCL} &
\textbf{Avg} \\
\midrule
Static &
56.0 & 65.5 & 21.0 & 52.0 & 48.6 \\
Greedy &
61.1 & 69.0 & 23.6 & 54.3 & 52.0 \\
Top-$k$ &
62.7 & 71.4 & 22.1 & 54.6 & 52.7 \\
\textcolor{black}{\textsc{EvoTool}} &
66.2 & 74.6 & 25.8 & 56.7 & 57.0 \\
\bottomrule
\end{tabular}
\end{adjustbox}
\vspace{2pt}
\caption{Ablation on population selection. All variants differ only in the parent selection strategy.}
\label{tab:ablation_selection}
\end{table}

\paragraph{Qualitative Analysis of Prompt Evolution.} As demonstrated in Appendix~\ref{app: P-Prompt} and Appendix~\ref{app: further case}, the initial Planner prompt provides only a generic instruction to decompose a user request into subgoals, leaving key ToolBench requirements under-specified. In contrast, EVOTOOL’s evolved Planner prompt encodes an explicit, executable planning policy and interface contract: it constrains planning to capabilities in tool index, forbids hallucinated identifiers or parameters, enforces atomic step-to-capability mapping, prioritizes early acquisition of required IDs, and standardizes stateful variable storage. It further introduces validation and lightweight fallback behaviors and requires a structured JSON plan with argument templates. These refinements demonstrate EvoTool’s effectiveness on planning heavy tasks, yielding prompts that better match ToolBench’s modes.





\section{Conclusion}
We introduced \textsc{EvoTool}, a self-evolving framework that optimizes modular tool-use policies through a gradient-free evolutionary paradigm to address the challenges of credit assignment under delayed supervision. By decomposing the agent policy into four distinct modules—Planner, Selector, Caller, and Synthesizer—\textsc{EvoTool} leverages trajectory-grounded blame attribution to localize failures, feedback-guided targeted mutation to execute precise natural-language updates, and diversity-aware population selection to prevent mode collapse. Empirical evaluations on four benchmarks, including ToolBench, RestBench, $\tau$-Bench, and BFCL, demonstrate that our approach consistently outperforms state-of-the-art baselines by more than 5 points on both GPT-4.1 and Qwen3-8B, while also achieving superior token efficiency and robust transferability across datasets.

\section*{Limitations}

While \textsc{EvoTool} demonstrates robust performance and efficiency, there are several avenues for further refinement. First, although the blame attribution and targeted mutation mechanisms significantly reduce token overhead compared to global optimization, the evolutionary process still necessitates iterative inference steps, which may introduce latency considerations for strictly real-time applications. Second, our current evaluation focuses on textual and API-based environments; extending this modular evolution paradigm to multi-modal tools or embodied agents remains an exciting direction for future research.

\bibliography{custom}

\clearpage
\appendix

\begin{algorithm*}[!t]
\caption{\textsc{EvoTool}: Self-Evolving Tool-use Policy Optimization}
\label{alg:evotool}
\begin{algorithmic}[1]
\Require \textbf{Inputs:} training pool $S_{train}$, selection set $S_{sel}$, frozen weights $W$, reward $R(\cdot)$, module set $\Pi$
\Require \textbf{Hyper-Parameters:} max generations $G$, mini-batch size $B$

\State \textbf{Initialize:} $\mathcal{P} \leftarrow \{\Theta^{(i)}\}_{i=1}^{N}$, where $\Theta=\{\theta_{plan},\theta_{sel},\theta_{call},\theta_{syn}\}$
\State Compute win frequency $\{w(\Theta)\}_{\Theta\in\mathcal{P}}$ on $S_{sel}$
\Comment{defined in Phase 3}

\For{$g = 1$ \textbf{to} $G$}

    \Phase{Phase 1: Trajectory Collection}
    \State Sample parent $\Theta_{p}\sim \mathcal{P}$ with $\Pr(\Theta_{p}=\Theta)\propto w(\Theta)$
    \State Sample mini-batch $\mathcal{B}\subset S_{train}$ with $|\mathcal{B}|=B$
    \State $\mathcal{E} \leftarrow \emptyset$
    \ForAll{$x\in\mathcal{B}$}
        \State $(\tau_x,\hat{y}_x)\leftarrow \textsc{Execute}(\pi_{\Theta_{p},W},x)$
        \State $r_x \leftarrow R(x,\hat{y}_x)$
        \State $e_x \leftarrow (x,\tau_x,\hat{y}_x,r_x)$
        \State $\mathcal{E} \leftarrow \mathcal{E}\ \Vert\ e_x$
    \EndFor
    \State $e \leftarrow \mathcal{E}$

    \Phase{Phase 2: Trajectory-Grounded Blame Attribution}
    \State $D \leftarrow \textsc{ExtractDiagnostics}(\tau)$
    \State $\{b_{\pi}(e)\}_{\pi\in\Pi} \leftarrow \textsc{BlamerLLM}(e,D)$
    \State $\pi^{*} \leftarrow \arg\max_{\pi\in\Pi} b_{\pi}(e)$
    \Phase{Phase 3: Feedback-Guided Targeted Mutation}
    \State $F(e,\pi^{*}) \leftarrow \textsc{MutatorLLM}(e,\pi^{*},D,\Theta_{p})$
    \State $\theta'_{\pi^{*}} \leftarrow \textsc{EditPrompt}(\theta_{p,\pi^{*}},F(e,\pi^{*}))$
    \State $\Theta_{child} \leftarrow \Theta_{p}$; \quad $\theta_{child,\pi^{*}}\leftarrow \theta'_{\pi^{*}}$

    \State $\bar{R}(\Theta;\mathcal{B}) \leftarrow \frac{1}{B}\sum_{x\in\mathcal{B}} R\!\left(x,\hat{y}_{\Theta}(x)\right)$
    \If{$\bar{R}(\Theta_{child};\mathcal{B}) > \bar{R}(\Theta_{p};\mathcal{B})$}
        \State $\mathcal{P} \leftarrow \mathcal{P}\cup\{\Theta_{child}\}$
    \EndIf

    \Phase{Phase 4: Diversity-Aware Population Selection (on $S_{sel}$)}
     \ForAll{$x\in S_{sel}$}
        \ForAll{$\Theta\in\mathcal{P}$}
            \State $(\tau_x,\hat{y}_x)\leftarrow \textsc{Execute}(\pi_{\Theta,W},x)$
            \State $r_x(\Theta) \leftarrow R(x,\hat{y}_x)$
        \EndFor
        \State $W(x)\leftarrow \arg\max_{\Theta\in\mathcal{P}} r_x(\Theta)$
    \EndFor
    \State $\mathcal{P} \leftarrow \{\Theta\in\mathcal{P}\mid \exists x\in S_{sel}\ \text{s.t.}\ \Theta = W(x)\}$
    \State $w(\Theta)\leftarrow \frac{1}{|S_{sel}|}\sum_{x\in S_{sel}}\mathbb{I}[\Theta = W(x)]\qquad \forall \Theta\in\mathcal{P}$

\EndFor

\State \Return $\Theta^{*}\in\arg\max_{\Theta\in\mathcal{P}}\ \frac{1}{|S_{sel}|}\sum_{x\in S_{sel}} R\!\left(x,\hat{y}_{\Theta}(x)\right)$
\end{algorithmic}
\end{algorithm*}

\section{Experiment Details}

\subsection{Algorithm Overview}
\label{app:algo}

Algorithm~\ref{alg:evotool} summarizes \textsc{EvoTool}’s self-evolving optimization loop. Starting from a population of module prompt specifications $\Theta=\{\theta_{\text{plan}},\theta_{\text{sel}},\theta_{\text{call}},\theta_{\text{syn}}\}$ with frozen LLM weights $W$, \textsc{EvoTool} iterates four phases: (1) collect execution trajectories and rewards on a mini-batch from $S_{\text{train}}$; (2) perform trajectory-grounded blame attribution to identify the most responsible module; (3) generate module-specific feedback and apply a targeted edit to mutate only that module, accepting the child if it improves mini-batch reward; (4) run diversity-aware population selection on a held-out set $S_{\text{sel}}$ by retaining policies that win at least one selection instance and updating sampling probabilities by win frequency. After $G$ generations, return the policy with the highest average reward on $S_{sel}$.}

\subsection{Datasets}
\label{app: datasets}

\begin{itemize}
  \item \textbf{ToolBench} \cite{qin2023toolllm} is a large-scale real-API tool-use benchmark from RapidAPI, where each instance pairs an instruction with tool/API documentation and requires executing multi-step API calls and producing a grounded final response. It covers 3,451 tools and 16,464 REST APIs, and includes three difficulty regimes: single-tool (G1), intra-category multi-tool (G2), and intra-collection multi-tool (G3). It mainly tests tool selection under large candidate sets, correct call ordering, and schema-valid argument construction.

  \item \textbf{RestBench} \cite{song2023restgpt} is a human-annotated sequential REST benchmark where each instruction is paired with a gold solution path. It contains two OpenAPI-based scenarios (TMDB with 51 APIs; Spotify with 40 APIs) and is designed to diagnose step-wise decomposition, correct endpoint ordering, and robustness across short dependent call chains.

  \item \textbf{$\tau$-Bench} \cite{yao2024tau} is a stateful customer-service dialogue benchmark where an agent follows domain policies while interacting with backend tools over a database; success is defined by reaching the correct final database outcome. It includes $\tau$-retail (1,000 orders) and $\tau$-airline (2,000 reservations). It emphasizes long-horizon planning, multi-turn state tracking, and policy-compliant tool use under ambiguous user requests.

  \item \textbf{BFCL} \cite{patil2025bfcl} is a function-calling benchmark that evaluates correct function selection and schema-valid arguments with deterministic checking. It includes 5,551 question--function--answer instances and a multi-turn suite with eight API suites and 1,000 queries. It emphasizes schema adherence, clarification when needed, and consistency across iterative tool-use turns.
\end{itemize}

\subsection{Baselines}
\label{app:baselines}

We compare \textsc{EvoTool} against three streams of tool-use policy design. For fairness, all methods use the same backbone LLM (frozen weights), the same tool sets/environments provided by each benchmark, and the same per-instance interaction budget. For any method that performs prompt search/optimization, we select the final prompt/policy using the same held-out selection set $S_{sel}$ protocol (best on $S_{sel}$), and report results on the benchmark’s official evaluation split.

\paragraph{Hand-crafted tool-use baselines.}
\begin{itemize}
  \item \textbf{ReAct} \cite{yao2022react} is a prompting framework that interleaves explicit reasoning traces with tool actions and uses tool observations to continue the trajectory. It targets the question of whether tightly coupling reasoning and acting improves reliability in interactive tasks and reduces hallucination through environment grounding.

  \item \textbf{Chain-of-Thought} \cite[CoT,][]{wei2022chain} is a prompting strategy that elicits intermediate reasoning steps (via exemplars) before producing the final answer. It targets the question of whether making multi-step reasoning explicit improves performance on tasks requiring compositional inference.

  \item \textbf{Plan-and-Solve} \cite{wang-etal-2023-plan} is a prompting framework that first generates an explicit high-level plan and then executes the plan to obtain the final answer. It targets the question of whether explicit decomposition reduces common multi-step reasoning failures (e.g., missing steps) compared to direct reasoning.
\end{itemize}

\paragraph{Monolithic tool-use policy optimization.}
\begin{itemize}
  \item \textbf{OPRO} \cite{yang2023large} is a derivative-free prompt optimization method that uses an LLM as an optimizer to iteratively propose improved prompts conditioned on past candidates and their scores. It targets the question of whether prompt search can be framed as black-box optimization in natural language without access to gradients or model parameters.

  \item \textbf{PromptBreeder} \cite{fernando2023promptbreeder} is an evolutionary framework that maintains a population of task prompts and mutates/selects them based on fitness, while also evolving the mutation prompts that generate new candidates. It targets the question of whether self-referential prompt evolution can automatically discover effective prompting strategies with minimal manual design.

  \item \textbf{EvoPrompt} \cite{guo2023connecting} is an evolutionary prompt optimizer that connects LLM generation to evolutionary operators (e.g., mutation/crossover) and selects prompts by development-set performance. It targets the question of whether combining evolutionary search with LLM-based candidate generation yields strong discrete prompt optimization across tasks and models.
\end{itemize}

\paragraph{Single-aspect tool-use policy optimization.}
\begin{itemize}
  \item \textbf{AdaPlanner} \cite{sun2023adaplanner} is a closed-loop LLM agent that generates an explicit plan and adaptively refines it using environment feedback during execution. It targets the question of whether feedback-driven plan refinement improves sequential decision-making as task horizons and complexity increase.

  \item \textbf{DRAFT} \cite{qu2024exploration} is a framework that iteratively refines tool documentation through self-driven tool interactions and feedback-based rewriting (explore--analyze--rewrite). It targets the question of whether improving tool documentation quality can directly improve LLM tool-use success and cross-model generalization.

  \item \textbf{EasyTool} \cite{yuan2025easytool} is a documentation transformation approach that condenses lengthy and heterogeneous tool descriptions into a unified, concise tool instruction/interface for agents. It targets the question of whether standardized, compact tool instructions improve tool selection and invocation while reducing context and noise.

  \item \textbf{AnyTool} \cite{du2024anytool} is a hierarchical tool-use agent that retrieves a small set of candidate APIs, solves the query with the selected candidates, and triggers self-reflection to retry when the initial solution is infeasible. It addresses the question of how to scale reliable tool use across very large API inventories through retrieval, structured solving, and self-correction.
\end{itemize}

\subsection{Meta Prompt for Blamer}

The Blamer meta prompt (Appendix \ref{app: Blamer Prompt}) defines a diagnostic judge that uses the task, full trajectory, module-level structured events, and a binary outcome to score Planner, Selector, Caller, and Synthesizer in 0 to 1 and select a single primary module; it prioritizes event evidence, validates with trajectory-grounded justification, and outputs scores, evidence, and a one-sentence diagnosis.

\begin{promptbox}[Blamer Meta Prompt]
\label{app: Blamer Prompt}
\small
\ttfamily
\raggedright
\footnotesize
\ttfamily
\raggedright
\setlength{\parindent}{0pt}
\setlength{\parskip}{0.6em}
\setlength{\emergencystretch}{20em} 

\vspace{1em}
\# ROLE \\
You are a diagnostic judge for a modular tool-using agent.

\vspace{1em}
\# GOAL \\
Given (i) a task, (ii) a full execution trajectory, (iii) structured events for each module in Planner,
Selector, Caller, and Synthesizer extracted from the trajectory, and (iv) an outcome signal
with either 0 (fail) or 1 (success), your task is to assign module-level blame to one of the four modules that is most responsible for the errors or suboptimality in the trajectory.

\vspace{1em}
\# ATTRIBUTION CRITERIA
\begin{itemize}[leftmargin=1.4em,label=-,nosep,topsep=0pt]
  \item Planner: missing or incorrect decomposition; incorrect ordering; dropped constraints or lost state.
  \item Selector: wrong tool choice; missing tool choice when necessary.
  \item Caller: schema or format violations; wrong parameters; malformed calls.
  \item Synthesizer: ungrounded final response; contradiction with tool outputs; missing integration of key observations.
\end{itemize}

\vspace{1em}
\# BLAME ASSIGNMENT RULES
\begin{itemize}[leftmargin=1.4em,label=-,nosep,topsep=0pt]
  \item Give each module a score in 0 to 1.
  \item Blame the most causal module that most directly caused failure or quality loss.
  \item Use the extracted events for each module first, then confirm with trajectory evidence.
  \item Prefer the earliest causal mistake. If multi causal, still pick one primary.
\end{itemize}

\vspace{1em}
\# OUTPUT FORMAT
Output plain text using following format:

1. Scores \\
planner \textless number\textgreater
selector \textless number\textgreater
caller \textless number\textgreater
synthesizer \textless number\textgreater
\vspace{0.5em}

2. Evidence \\
Provide evidence for each module. Each line must include information from the extracted events
and a short reason grounded in the trajectory.
\vspace{0.5em}

3. One sentence diagnosis \\
Write one sentence explaining why the primary module is blamed.
\end{promptbox}

\subsection{Meta Prompt for Mutator}

\begin{promptbox}[Mutator Meta Prompt]
\label{app: Mutator Prompt}
\small
\ttfamily
\raggedright
\footnotesize
\ttfamily
\raggedright
\setlength{\parindent}{0pt}
\setlength{\parskip}{0.7em}
\setlength{\emergencystretch}{2em} 

\# ROLE\\
You are a targeted prompt editor for exactly one module of a modular tool-using agent.

\vspace{1em}
\# GOAL \\
Given (i) a target module chosen from Planner, Selector, Caller, and Synthesizer, (ii) the current specification of that module, (iii) a failure episode packet containing the task input \texttt{x}, the module-local trajectory slice (the target module's outputs plus nearby context), the final outcome and verifier feedback, and (iv) the blamer's rationale and blame scores, produce a single minimal and general edit to the selected module that addresses the diagnosed failure mode while preserving the module's interface contract and output format.
\vspace{1em}

\# EDITING RULES\\
- Edit only target module specification; do not modify other modules.\\
- Do not add new tools or environments.\\
- Ground the edit in the trajectory \\
- Make the smallest change that fixes the error or suboptimality.

\vspace{1em}
\# HEURISTIC EDIT PATTERNS\\
- Schema/format error $\rightarrow$ add argument checklist, schema verification.\\
- Wrong tool selection $\rightarrow$ add decision rubric mapping subgoals to tools.\\
- Planning error $\rightarrow$ add explicit subgoals, state fields, ordering constraints, prerequisite checks.\\
- Ungrounded synthesis $\rightarrow$ require attribution to tool outputs, prohibit unsupported facts.

\vspace{1em}
\# OUTPUT FORMAT\\
Output plain text with the following sections in this order:

\vspace{0.5em}
1. Target module\\
\textless planner or selector or caller or synthesizer\textgreater

\vspace{0.5em}
2. Diagnosed error mode\\
\textless 1--2 sentences describing the failure mode grounded in the trajectory \textgreater

\vspace{0.5em}
3. Minimal edit summary\\
\textless 1--2 short sentences describing the minimal change and why\textgreater

\vspace{0.5em}
4. Revised target module spec\\
\textless updated specification text for the target module only\textgreater
\end{promptbox}

The Mutator meta prompt (Appendix \ref{app: Mutator Prompt}) defines a targeted editor that, given the Blamer-selected module, its current specification, a failure episode packet, and the Blamer rationale and scores, produces one minimal general edit to the selected module while preserving its interface and format; it restricts edits to the target module, requires trajectory grounding, and outputs the diagnosed error mode, edit summary, and revised specification.

\subsection{Implementation Details}
\label{app: implementation details}

We utilize two distinct base Large Language Models to evaluate the generalizability of \textsc{EvoTool}: the commercial GPT-4.1\footnote{\url{https://openai.com/index/gpt-4-1/}} \citep{openai2025gpt41} and the open-source Qwen3-8B\footnote{\url{https://huggingface.co/Qwen/Qwen3-8B}} \citep{yang2025qwen3}. For fairness, across all baselines we use the same backbone LLM, the benchmark-provided tool environments, and the same per-instance interaction budget; unless a method is purely hand-crafted, its final prompt is selected by best performance on $S_{\text{sel}}$ and evaluated on the benchmark’s official split. We set the maximum generation budget $G$ to 8 iterations. The mini-batch size $B$ is set to be 3.

\subsection{Initial Prompt Setup}
We list the initial module prompts.

\begin{promptbox}[Initial Planner Prompt]
\label{app: P-Prompt}
\small
\ttfamily
\raggedright
\footnotesize
\ttfamily
\raggedright
\setlength{\parindent}{0pt}
\setlength{\parskip}{0.7em}
\setlength{\emergencystretch}{2em} 

You are a planning agent. Your task is to decompose the user's complex instruction into a sequential list of clear, executable subgoals
\end{promptbox}

\begin{promptbox}[Initial Selector Prompt]
\label{app: S-Prompt}
\small
\ttfamily
\raggedright
\footnotesize
\ttfamily
\raggedright
\setlength{\parindent}{0pt}
\setlength{\parskip}{0.7em}
\setlength{\emergencystretch}{2em} 

You are a tool selection agent. Given the current subgoal and the list of available tools, select the most appropriate tool.
\end{promptbox}

\begin{promptbox}[Initial Caller Prompt]
\label{app: C-Prompt}
\small
\ttfamily
\raggedright
\footnotesize
\ttfamily
\raggedright
\setlength{\parindent}{0pt}
\setlength{\parskip}{0.7em}
\setlength{\emergencystretch}{2em} 

You are a tool calling agent. Given the selected tool and its documentation, generate the specific arguments required to execute it.
\end{promptbox}

\begin{promptbox}[Initial Synthesizer Prompt]
\label{app: Sy-Prompt}
\small
\ttfamily
\raggedright
\footnotesize
\ttfamily
\raggedright
\setlength{\parindent}{0pt}
\setlength{\parskip}{0.7em}
\setlength{\emergencystretch}{2em} 

You are a synthesis agent. Review the user's original query and the history of tool executions, then synthesize this information to provide the answer.
\end{promptbox}

\subsection{Further Cases}
\label{app: further case}

\begin{figure*}[!t]
\centering
\begin{promptbox}[Final Planner Prompt]
\label{app:P-Prompt}
\footnotesize
\ttfamily
\raggedright
\setlength{\parindent}{0pt}
\setlength{\parskip}{0.5em}
\setlength{\emergencystretch}{2em}
You are the \textbf{PLANNER} module in a modular tool-using agent for ToolBench (RapidAPI-style REST tools).

\textbf{Mission:} Convert \texttt{USER\_TASK} into a \emph{minimal, fully executable} tool plan grounded in \texttt{TOOL\_INDEX} and \texttt{STATE}. You must plan actions only (no tool calls, no final answers).

\textbf{Inputs:} \texttt{USER\_TASK} (constraints/preferences), \texttt{TOOL\_INDEX} (tool names + brief capabilities), \texttt{STATE} (cached variables from earlier steps).

\textbf{Rules (ToolBench-critical).}
1) \textbf{Grounding:} reference only tool names/capabilities that appear in \texttt{TOOL\_INDEX}. If a needed capability is absent, plan the best workaround and record the gap in \texttt{notes}.  
2) \textbf{No hallucination:} never invent IDs, parameters, formats, or facts. All non-trivial values must come from \texttt{STATE} or a planned tool step.  
3) \textbf{Atomicity:} each step maps to exactly one concrete capability (search/list/details/lookup/geocode/convert/compute/filter). Avoid vague steps (``analyze'', ``research'').  
4) \textbf{Dependency-first:} identify required identifiers/keys (\texttt{item\_id}, \texttt{place\_id}, \texttt{user\_id}, \texttt{lat/lon}, etc.) and obtain them as early as possible.  
5) \textbf{Minimality:} use the fewest steps that guarantee correctness; prefer a single filtered search call over multiple redundant calls.  
6) \textbf{State discipline:} reuse \texttt{STATE}; explicitly specify what to store after each step (variable names must be stable and descriptive).  
7) \textbf{Validation:} include a step to verify critical constraints (time window, region, availability, price bounds, unit/currency).  
8) \textbf{Fallback:} include at most one lightweight fallback per step for \texttt{empty\_result|error|missing\_id|format\_issue} (e.g., broaden query, relax filter, alternate tool).

\textbf{Procedure.}
A) Parse \texttt{USER\_TASK}: separate must-haves vs preferences; normalize time/units/location into explicit variables (ISO dates, currency code, radius km).  
B) Scan \texttt{TOOL\_INDEX} and choose the smallest relevant tool set (prefer 1--2 tools).  
C) Plan in executable order: (i) disambiguating search/list $\rightarrow$ (ii) select candidate(s) $\rightarrow$ (iii) details/lookup $\rightarrow$ (iv) validate constraints $\rightarrow$ (v) produce structured intermediate results.  
D) If the task is underspecified, do \emph{tool-based disambiguation} (top-$k$ candidates + details) rather than asking the user. Record assumptions to verify.

\textbf{Variable conventions (STATE).}
Use explicit names: \texttt{query}, \texttt{candidates}, \texttt{selected\_id}, \texttt{details}, \texttt{lat}, \texttt{lon}, \texttt{start\_date}, \texttt{end\_date}, \texttt{price\_min}, \texttt{price\_max}. Store candidate arrays as \texttt{candidates[i].id/name/reason}. Avoid overwriting high-value vars.

\textbf{Output: JSON only (exactly one object).} Use this schema:
\begin{verbatim}
{
  "plan":[
    {"step_id":1,
     "subgoal":"...",
     "tool_name":"(EXACT from TOOL_INDEX)",
     "tool_capability_hint":"search|list|details|lookup|geocode|convert|compute|filter",
     "arguments_template":{"param":"..."},
     "required_inputs":["STATE.x","USER_TASK.y"],
     "expected_outputs":["id","fields"],
     "store_as":{"STATE.var":"from_output.field"},
     "success_criteria":["non-empty id/fields; constraints satisfied"],
     "fallback":{"when":"empty_result|error|missing_id|format_issue",
                 "action":"broaden/relax/alternate-tool/retry-format",
                 "arguments_template":{"changed_param":"..."}},
     "depends_on":[]
    }
  ],
  "notes":["must-have constraints","preferences","assumptions to verify","tool gaps (if any)"]
}
\end{verbatim}

\textbf{Now produce the plan. Output JSON only.}
\end{promptbox}
\end{figure*}

\begin{figure*}[!t]
\centering
\begin{promptbox}[Final Selector Prompt]
\label{app:S-Prompt}
\footnotesize
\ttfamily
\raggedright
\setlength{\parindent}{0pt}
\setlength{\parskip}{0.5em}
\setlength{\emergencystretch}{2em}
You are the \textbf{SELECTOR} module in a modular tool-using agent for ToolBench (RapidAPI-style REST tools).

\textbf{Mission:} Given \texttt{USER\_TASK}, \texttt{TOOL\_INDEX}, and current \texttt{STATE}, choose the \emph{single best next tool action} to execute, including a concrete tool name and arguments. You do not answer the user; you only select the next tool call.

\textbf{Inputs.}
- \texttt{USER\_TASK}: the user request and constraints.
- \texttt{TOOL\_INDEX}: available tools/APIs (names + descriptions).
- \texttt{STATE}: accumulated variables/results from prior steps (may include a plan, candidates, IDs, partial details).
- (Optional) \texttt{PLAN}: a JSON plan produced by the PLANNER (if present, follow it).

\textbf{Selector rules (ToolBench-critical).}
1) \textbf{Grounding:} select only a \texttt{tool\_name} that appears in \texttt{TOOL\_INDEX}.  
2) \textbf{Argument fidelity:} use only argument keys that are supported by the chosen tool; if uncertain, prefer the simplest valid call (often search/list) and record uncertainty in \texttt{rationale}.  
3) \textbf{State-first:} reuse IDs/values already in \texttt{STATE}; never re-search if an ID is available.  
4) \textbf{Dependency correctness:} if downstream steps require an ID/key, prioritize actions that obtain it earliest.  
5) \textbf{Minimal progress:} choose the next action that maximizes information gain and reduces uncertainty (disambiguate $\rightarrow$ get details $\rightarrow$ validate constraints).  
6) \textbf{No hallucination:} do not fabricate tool outputs; do not assume availability/price/details without tool evidence.  
7) \textbf{Failure-awareness:} if the previous tool call failed or returned empty, apply exactly one fallback adjustment (broaden query, relax filters, alternate tool) consistent with the plan/notes.

\textbf{Decision procedure.}
A) Identify the next unmet requirement from \texttt{PLAN} (if present); otherwise infer the highest-priority missing variable for completing the task (often \texttt{selected\_id}, \texttt{lat/lon}, availability, price).  
B) Choose the tool/capability that directly produces that requirement.  
C) Construct \texttt{arguments} using \texttt{STATE} values. If required fields are missing, perform a disambiguating search/list (top-$k$) rather than guessing.  
D) Ensure arguments respect must-have constraints (time, region, budget, units).  
E) If multiple candidates exist, prefer retrieving details for the top candidate(s) rather than expanding search breadth.

\textbf{Output format (JSON only).} Return exactly one JSON object:
\begin{verbatim}
{
  "tool_name": "EXACT tool name from TOOL_INDEX",
  "tool_capability_hint": "search|list|details|lookup|geocode|convert|compute|filter",
  "arguments": { "param_a": "value", "param_b": "value" },
  "store_as": { "STATE.var": "from_tool_response.field(s)" },
  "stop_condition": "What evidence would indicate this step is sufficient 
  (e.g.4 have item_id and details)",
  "fallback": {
    "when": "empty_result|error|missing_id|format_issue",
    "tool_name": "alternate tool name (optional, must be in TOOL_INDEX)",
    "arguments": { "changed_param": "..." }
  },
  "rationale": [
    "1-3 short bullets: why this is the best next action given PLAN/STATE and constraints"
  ]
}
\end{verbatim}

\textbf{Final instruction.} Output JSON only. Do not include any explanation outside the JSON.
\end{promptbox}
\end{figure*}

\begin{figure*}[t]
\centering
\begin{promptbox}[Final Caller Prompt]
\label{app:C-Prompt}
\footnotesize
\ttfamily
\raggedright
\setlength{\parindent}{0pt}
\setlength{\parskip}{0.5em}
\setlength{\emergencystretch}{2em}
You are the \textbf{CALLER} module in a modular tool-using agent for ToolBench (RapidAPI-style REST tools).

\textbf{Mission:} Execute exactly one tool/API call specified by the SELECTOR. Your job is to (i) validate and format the call, (ii) make the call, and (iii) write back a clean, minimal \texttt{STATE} update based only on the tool response. You do not answer the user.

\textbf{Inputs.}
- \texttt{USER\_TASK}
- \texttt{TOOL\_INDEX} (tool names + brief descriptions)
- \texttt{STATE} (current working memory)
- \texttt{NEXT\_ACTION}: a JSON object from SELECTOR with \texttt{tool\_name}, \texttt{arguments}, \texttt{store\_as}, and optional \texttt{fallback}.

\textbf{Caller rules (ToolBench-critical).}
1) \textbf{One call only:} execute exactly one tool call per turn (either the primary call, or the fallback if triggered).  
2) \textbf{Grounding:} the \texttt{tool\_name} must exist in \texttt{TOOL\_INDEX}. If not, do not guess; return an error in output.  
3) \textbf{Argument validation:} pass only keys supported by the chosen tool. If unsupported keys exist, drop them (do not invent replacements). If required keys are missing, trigger fallback when provided; otherwise return an error.  
4) \textbf{No hallucination:} never fabricate outputs. All stored values must be directly extracted from the tool response.  
5) \textbf{Safe parsing:} handle empty, partial, or nested responses; extract IDs/fields defensively.  
6) \textbf{State updates only:} store outputs exactly as \texttt{store\_as} indicates, plus minimal metadata (e.g., \texttt{last\_tool}, \texttt{last\_status}). Do not overwrite high-value variables unless explicitly requested by \texttt{store\_as}.  
7) \textbf{Failure handling:} if the primary call returns empty/error and a fallback exists, execute the fallback instead (still only one call total). Record what happened.

\textbf{Output format (JSON only).} Return exactly one JSON object:
\begin{verbatim}
{
  "called": {
    "tool_name": "...",
    "arguments": { "k": "v" }
  },
  "status": "success|empty|error",
  "raw_summary": "1-2 short sentences summarizing what the tool returned (no invented facts)",
  "extracted": { "key_fields": "values actually present in response" },
  "state_update": { "STATE.var": "stored value(s) per store_as", 
  "STATE.last_tool": "...", "STATE.last_status": "..." },
  "error": { "type": "...", "message": "..." }
}
\end{verbatim}

\textbf{Notes.}
- \texttt{raw\_summary} must not include any values that are not present in the response.  
- \texttt{error} must be \texttt{null} when \texttt{status="success"}.

\textbf{Final instruction.} Output JSON only. Do not include any explanation outside the JSON.
\end{promptbox}
\end{figure*}

\begin{figure*}[t]
\centering
\begin{promptbox}[Final Synthesizer Prompt]
\label{app:Y-Prompt}
\footnotesize
\ttfamily
\raggedright
\setlength{\parindent}{0pt}
\setlength{\parskip}{0.5em}
\setlength{\emergencystretch}{2em}
You are the \textbf{SYNTHESIZER} module in a modular tool-using agent for ToolBench (RapidAPI-style REST tools).

\textbf{Mission:} Produce the final user-facing answer using only verified information in \texttt{STATE} (i.e., tool outputs from CALLER) and the original \texttt{USER\_TASK}. Do \emph{not} call tools. Do \emph{not} invent facts.

\textbf{Inputs.}
- \texttt{USER\_TASK}: the user request, constraints, preferences.
- \texttt{STATE}: accumulated results from executed tools (may include candidates, selected IDs, details, computed values, and intermediate tables).
- (Optional) \texttt{PLAN}: high-level plan for traceability (do not restate verbatim unless asked).

\textbf{Synthesis rules (ToolBench-critical).}
1) \textbf{Ground truth:} every factual claim must be supported by \texttt{STATE}. If a required fact is missing, explicitly say what is missing and what tool evidence would be needed (but do not call tools).  
2) \textbf{Task alignment:} satisfy must-have constraints first; then optimize for preferences.  
3) \textbf{Consistency checks:} verify that IDs, dates, locations, prices, and units are consistent across stored fields; if conflicts exist, surface them and choose the most reliable field (prefer ``details'' endpoints over ``search'' snippets).  
4) \textbf{No leakage:} do not mention internal modules (PLANNER/SELECTOR/CALLER), tool invocation mechanics, or private reasoning.  
5) \textbf{Helpful structure:} present results in a concise, scannable format (bullets or short sections). Include options/rankings when relevant.  
6) \textbf{Uncertainty handling:} if multiple candidates remain, present top options with brief evidence from \texttt{STATE} and a recommendation criterion.  
7) \textbf{Safety against overreach:} do not provide speculative instructions or claims beyond available evidence; keep language precise (e.g., ``The API response shows...").

\textbf{Output requirements.}
- Write in the user's language.
- Prefer concrete values (dates, amounts, names) only if present in \texttt{STATE}.
- If the user asked for a specific format (table/JSON/list), follow it.

\textbf{Output format (plain text only).}
Return the final response as normal text for the user (not JSON). Do not include citations unless the user explicitly requested them.

\textbf{Final instruction.} Using only \texttt{USER\_TASK} and \texttt{STATE}, write the final answer now.
\end{promptbox}
\end{figure*}

\begin{table*}[htbp]
\centering
\small
\resizebox{\textwidth}{!}{%
\begin{tabular}{@{}p{0.15\linewidth} p{0.25\linewidth} p{0.52\linewidth}@{}}
\toprule
\textbf{Stage} & \textbf{Artifact} & \textbf{Content} \\
\midrule

Benchmark &
Task instruction &
I’m planning a hiking trip... (omitted for brevity) \newline
Fetch nearby places for 39.5501 N, 105.7821 W using \textbf{TrueWay Places}. \newline
Get geocode for `987 Oak Street' using \textbf{Census Bureau}. \\

Benchmark &
Ground solution &
R1: \texttt{findplacesnearby} (TrueWay) \newline
R2: \texttt{geocoding} (Census Bureau) \newline
R3: Provide Answer \\

\midrule

Episode 100 (failure) &
Selector Spec v0 &
\texttt{[Selector Spec v0 — minimal]} \newline
- Choose next tool based on semantic relevance. \newline
- Prefer tools that directly answer the question. \newline
- Pick most common/general tool if unsure. \\

Episode 100 &
Incorrect selection &
\texttt{Tool=Weather; API=getForecast;} \newline
\texttt{Rationale=Geocoding seems related to location context.} \\

Episode 100 &
Feedback &
\texttt{Error: required tool not used.} \newline
\texttt{Expected: Geocoder - US Census Bureau.} \newline
\texttt{Observed: Weather.} \\

\midrule

EvoTool &
\textbf{Evolved prompt} &
\texttt{[Selector Spec v1]} \newline
\texttt{HARD CONSTRAINTS:} \newline
1) If user explicitly requires tools, you MUST select them. \newline
2) Preserve required tool order. \newline
\texttt{SELECTION HEURISTIC:} \newline
A) Identify missing variable (subgoal). \newline
B) Pick required tool producing that variable. \\

\midrule

Episode 101 (success) &
Selection (round 1) &
\texttt{Tool=TrueWay Places; API=findplacesnearby;} \newline
\texttt{Rationale=Fetch nearby lodging.} \\

Episode 101 &
Tool call 1 &
\texttt{Tool: TrueWay Places} \newline
\texttt{Args: \{lat: 39.55, long: -105.78, ...\}} \\

Episode 101 &
Tool output 1 &
\texttt{[ \{"name":"Place\_1", ...\}, ... ]} \\

Episode 101 &
Selection (round 2) &
\texttt{Tool=Census Bureau; API=geocoding;} \newline
\texttt{Rationale=Get lat/lon for address.} \\

Episode 101 &
Tool call 2 &
\texttt{Tool: Geocoder - US Census Bureau} \newline
\texttt{Args: \{"address":"987 Oak Street"\}} \\

Episode 101 &
Final answer &
Nearby places: Place\_1, Place\_2... \newline
Geocode: (LAT, LON) \\

\bottomrule
\end{tabular}
}
\caption{ToolBench Case}
\label{tab:evotool-toolbench-case}
\end{table*}

\end{document}